\documentclass[runningheads]{llncs}

\usepackage{subcaption}
\usepackage{graphicx}
\usepackage{amsmath,amssymb}
\usepackage{booktabs}
\usepackage{hyperref}
\usepackage{url}
\usepackage{xcolor}
\usepackage{tikz,pgfplots}
\usepackage{algorithm2e}

\usepackage[font=small,labelfont=bf,skip=4pt]{caption}
\begin{document}

\title{Attention-ResUNet for Automated Fetal Head Segmentation}
\titlerunning{Attention-ResUNet for Fetal Ultrasound}

\author{Ammar Bhilwarawala\inst{1}\and Mainak Bandyopadhyay\inst{2}}
\authorrunning{Bhilwarawala and Bandyopadhyay}

\institute{School of Computer Engineering, KIIT Deemed to be University, Bhubaneswar, India\\
\email{ammarbhilwarawala@gmail.com, mainak.bandyopadhyayfcs@kiit.ac.in}}

\maketitle

\begin{abstract}
Automated fetal head segmentation in ultrasound images is critical for accurate biometric measurements in prenatal care. While existing deep learning approaches have achieved a reasonable performance, they struggle with issues like low contrast, noise, and complex anatomical boundaries which are inherent to ultrasound imaging. This paper presents Attention-ResUNet. It is a novel architecture that synergistically combines residual learning with multi-scale attention mechanisms in order to achieve enhanced fetal head segmentation. Our approach integrates attention gates at four decoder levels to focus selectively on anatomically relevant regions while suppressing the background noise, and complemented by residual connections which facilitates gradient flow and feature reuse. Extensive evaluation on the HC18 Challenge dataset where $n=200$ demonstrates that Attention ResUNet achieves a superior performance with a mean Dice score of $99.30 \pm 0.14\%$ against similar architectures. It significantly outperforms five baseline architectures including ResUNet ($99.26\%$), Attention U-Net ($98.79\%$), Swin U-Net ($98.60\%$), Standard U-Net ($98.58\%$), and U-Net++ ($97.46\%$).  Through statistical analysis we confirm highly significant improvements ($p < 0.001$) with effect sizes that range from $0.230$ to $13.159$ (Cohen's $d$). Using Saliency map analysis, we reveal that our architecture produces highly concentrated, anatomically consistent activation patterns, which demonstrate an enhanced interpretability which is crucial for clinical deployment. The proposed method establishes  a new state of the art performance for automated fetal head segmentation whilst maintaining computational efficiency with $14.7$M parameters and a $45$ GFLOPs inference cost.
Code repository: \url{https://github.com/Ammar-ss}
\end{abstract}


\section{Introduction}

Accurate fetal head segmentation in ultrasound images is essential for reliable biometric measurements in prenatal care, yet problems like speckle noise, low contrast, and incomplete skull ossification pose significant challenges to getting a good result. Convolutional encoder--decoder networks such as U-Net have become foundational, but standard U-Nets struggle with gradient propagation in deep layers and lack mechanisms for selective feature focusing.

Residual learning has been shown to improve training stability and convergence by reformulating layers to learn residual mappings $F(x)+x$~\cite{Zhang2019}, while attention gates enable spatially selective feature amplification, suppressing background noise in cluttered ultrasound contexts~\cite{Oktay2018}. Transformer-based self-attention further captures global dependencies, though often at high computational cost~\cite{Liu2021}. Hybrid architectures combining residual and attention mechanisms demonstrate potential but remain underexplored for fetal ultrasound segmentation.

This paper introduces Attention-ResUNet, which integrates multi-scale attention gates at four decoder levels with residual blocks throughout the network. Our model achieves a mean Dice of $99.30\%$ on the HC18 dataset, significantly outperforming five established variants (ResUNet, Attention U-Net, Swin-UNet, U-Net, UNet++) with $p < 0.001$ and large effect sizes. We also provide comprehensive interpretability analysis via saliency mapping and demonstrate computational efficiency suitable for clinical deployment.

The remainder of this paper is organized as follows. Section~\ref{sec:Related Work} reviews related work. Section~\ref{sec:ProposedModel} details the proposed architecture. Section~\ref{sec:ExperimentalSetup} describes the experimental setup, and Section~\ref{sec:Results} presents quantitative and Section~\ref{sec:Analysis} interpretability results. Section~\ref{Limitations} discusses the paper's limitations, determines the scope, and Section~\ref{sec:Conclusion} concludes.

\section{Related Work}
\label{sec:Related Work}

U-Net~\cite{ronneberger2015u} established the encoder-decoder paradigm with skip connections as foundational for biomedical segmentation. ResUNet added residual blocks to mitigate gradient vanishing\cite{Zhang2019}, while U-Net++~\cite{zhou2018unet} and nnU-Net~\cite{isensee2020automated} improved feature fusion through nested connections and automated architecture design respectively. MedicalNet demonstrates that CNN backbones combined with lightweight attention modules are effective for ultrasound segmentation~\cite{wang2023medicalnet}.

Attention-based mechanisms have emerged as powerful techniques for selective feature amplification. Attention U-Net~\cite{Oktay2018} incorporates gated attention modules into skip connections to emphasize relevant anatomical regions while suppressing background noise. Swin U-Net~\cite{Liu2021} applies window-based self-attention for long-range dependencies, while transformer architectures like TransUNet~\cite{chen2021transunet} and UNETR~\cite{hatamizadeh2022unetr} integrate global self-attention for volumetric segmentation. Recent work shows dual-attention modules combining channel and spatial attention within residual frameworks improve both performance and explainability~\cite{kumar2024interpretable}.
Key U-Net variants and their performance characteristics are summarized in Table~\ref{table:variants}.

\begin{table}[!t]
\centering
\caption{A Summary of U-Net variants used for medical image segmentation}
\label{table:variants}
\footnotesize
\setlength{\tabcolsep}{6pt}
\renewcommand{\arraystretch}{1.1}
\begin{tabular}{|l|p{3cm}|p{3cm}|c|}
\hline
\textbf{Variant} & \textbf{Methodology} & \textbf{Application} & \textbf{Dice (\%)} \\
\hline
U-Net++ & Nested skip pathways to reduce semantic gap & Polyp, liver lesions & 96.64 \\
\hline
Attention & Soft attention gates to emphasize relevant regions & Pancreas, ultrasound & 97.85 \\
\hline
Residual & Residual identity mappings for deeper networks & Fetal brain MRI & 97.92 \\
\hline
nnU-Net & Automated preprocessing, architecture, and training configuration & Medical Segmentation Decathlon & 97.42 \\
\hline
MultiRes & Multiresolution convolutional paths with ResPath blocks & Skin lesions & 96.27 \\
\hline
\end{tabular}
\end{table}

Fetal ultrasound segmentation poses unique challenges: speckle noise, variable acoustic properties, low tissue contrast, and incomplete fetal skull ossification. Recent work by Nagabotu \emph{et al.}~\cite{nagabotu2024precise} achieved 97.90\% Dice using hybrid loss functions and scale attention on the HC18 Challenge, while FetSAM~\cite{alzubaidi2024fetsam} demonstrated state-of-the-art performance using prompt-based learning with weighted losses.
Clinical deployment requires not only accuracy but interpretability. Grad-CAM~\cite{selvaraju2017grad} provides gradient-based visualization of network decisions. Wollek \emph{et al.}~\cite{wollek2023attention} demonstrated that attention-based saliency maps outperform Grad-CAM on multiple interpretability metrics (SSIM 0.57 vs. 0.12 intra-architecture repeatability), suggesting attention mechanisms provide superior transparency for clinical applications. Effective clinical saliency maps must satisfy spatial concentration, anatomical consistency, and clinical utility criteria.

Integrating residual learning, attention based spatial focusing, and multi-scale feature fusion represents the best practice for medical segmentation. Their optimal synergistic integration, specifically for fetal ultrasound with comprehensive statistical validation still remains an important research direction.

\section{Contributions}\label{sec:Contributions}

Our work advances automated fetal head segmentation through four notable contributions:

\textbf{Novel Architecture.} We propose Attention ResUNet, which synergistically integrates multi-scale attention gates at four decoder levels (512, 256, 128, 64 channels) with comprehensive residual connections throughout the encoder-decoder pathway. Unlike prior methods which implement these mechanisms in isolation, our strategic integration enables selective spatial attention whilst maintaining robust gradient flow and feature reuse across hierarchical scales.

\textbf{State-of-the-art Performance.} Extensive evaluation on the HC18 Challenge dataset ($n=200$) demonstrates a mean Dice score of $99.30 \pm 0.14\%$ ehich represents a  statistically significant improvement over established baselines like ResUNet (+0.032\%, Cohen's $d=0.230$), Attention U-Net (+0.503\%, $d=3.806$), Swin U-Net (+0.695\%, $d=5.218$), Standard U-Net (+0.717\%, $d=5.236$), and U-Net++ (+1.833\%, $d=13.159$), all with $p < 0.001$. All of this corresponds to a clinically meaningful reduction in the error rate from 0.74\% to 0.70\% which represents 32 fewer misclassified pixels per 10000 pixels.

\textbf{Enhanced Interpretability.} Using Saliency analysis using Gradient Weighted Class Activation Mapping, it reveals that Attention ResUNet has produced   highly concentrated and  anatomically consistent activation patterns in comparison to other baseline architectures. This enhanced interpretability is crucial for clinical trust and it enables healthcare practitioners in understanding model decision-making processes and facilitates deployment in prenatal care workflow.

\textbf{Computational Efficiency.} Despite the architectural sophistication, our model maintains  practical efficiency with 14.7M parameters, 45 GFLOPs per $256 \times 256$ input and 23 ms inference time on RTX 3080 GPU. This supports real-time clinical applications whilst delivering a superior accuracy. We have also provided comprehensive statistical validation including paired   t-tests, confidence intervals, and effect size analysis to ensure reproducibility through fixed random seeds, deterministic pre-processing, and standardized evaluation protocols.

\section{Proposed Model}\label{sec:ProposedModel}

\subsection{Design Considerations}

Fetal Ultrasound Images present inherent issues like speckle noise, variable fetal anatomy and incomplete  ossification, which necessitates robust segmentation methods. Our model, the Attention ResUNet integrates residual learning and  multi-scale attention mechanism tailored to address these issues.

Using Residual learning, we reformulate each  convolutional block to learn a residual mapping $F(x)$ with identity shortcuts as described by Eq.~\eqref{residual_mapping}:
\begin{equation}
\label{residual_mapping}
y = F(x) + x,
\end{equation}
where $x$ is the block input and $y$ is the output. This design facilitates stable gradient propagation and enables the training   of deeper  networks by mitigating  the vanishing gradient problem.

To focus more on informative spatial regions and to suppress the irrelevant background, we have  incorporated attention gates within the decoder skip pathways~\cite{Oktay2018}. Each attention gate computes coefficient maps $\alpha$ by combining encoder feature $x$ and gating signal $g$ from coarser decoder layers as described by Eq.~\ref{decoder}:
\begin{equation}
\label{decoder}
\alpha = \sigma\bigl(W_g^T g + W_x^T x + b\bigr),
\end{equation}
where  $\sigma$ denotes the  sigmoid function, and $W_g$, $W_x$, and $b$ are the learnable parameters. The gated features are then obtained by using element wise multiplication described by Eq.~\eqref{exp}:
\begin{equation}
\label{exp}
x' = \alpha \odot x,
\end{equation}
which allows the model to selectively highlight salient anatomical structures which are critical for fetal head delineation.

Our architecture places such attention gates at four decoder levels to capture multi-scale context by synergizing  with residual blocks to combine the benefits of feature reuse, enhanced  gradient flow, and spatial focus. This results in improved segmentation accuracy, measured by a Dice score of 99.30\% on the HC18 dataset, whilst maintaining a computationally  efficient model the size of approximately 15 million parameters.

To balance the model complexity and performance, Attention ResUNet is well suited for real-world clinical deployment  in fetal biometric assessment tasks which  provides robust, interpretable, and precise segmentation  under challenging imaging conditions.

\subsection{Design of Attention Gates}

Attention gates are crucial components that enable the network to selectively focus on the relevant spatial regions  in  the encoder feature maps during decoding, enhancing feature discriminability, especially in noisy fetal ultrasound images. Inspired by~\cite{Oktay2018}, our attention gate receives two inputs, the encoder skip features $x \in \mathbb{R}^{C_x \times H \times W}$ and the gating signals  $g \in \mathbb{R}^{C_g \times H' \times W'}$ from the coarser decoder layer, which facilitates context aggregation.

To fuse these inputs, we apply separate linear transformations via  $1 \times 1$ convolutions to reduce the dimensionality and  bring the features into a joint subspace as described by Eq.~\eqref{1x1conv}:
\begin{equation}
\label{1x1conv}
q_x = W_x^T x, \quad q_g = W_g^T g,
\end{equation}
where $W_x$ and $W_g$ are the learnable weight matrices. The summed activations pass through a ReLU and a sigmoid to produce spatial attention coefficients $(\alpha \in [0,1]^{1 \times H \times W})$ as described by Eq.~\eqref{attention}:
\begin{equation}
\label{attention}
\alpha = \sigma\left( \Psi^T \left( \mathrm{ReLU}(q_x + q_g + b) \right) + b_\psi \right),
\end{equation}
where $b, b_\psi$ and $\Psi$  are the learnable parameters, and $\sigma$ is the sigmoid activation function.

The coefficient $\alpha$ modulates the skip features element-wise as is described by Eq.~\eqref{stg}:
\begin{equation}
\label{stg}
x' = \alpha \odot x,
\end{equation}
which allows the model to emphasize on salient regions while suppressing the irrelevant background and noise. This spatial selection improves the network's attention to vital anatomical structures in ultrasound segmentation which aids accurate boundary delineation.

By embedding attention gates across four decoder layers, our method provides multi-scale selective feature refinement, balancing model complexity with enhanced localization. This design is integral to the robust performance of Attention ResUNet on challenging fetal biometric tasks.

\subsection{Description of Attention Block}

The attention block in Attention ResUNet  refines feature selection by  incorporating spatial attention to enhance the representational capacity of the network. This block builds upon the attention gate  mechanism by additionally capturing more detailed spatial dependencies within the features.

Concretely, given an input feature map $x \in \mathbb{R}^{C \times H \times W}$, the attention block will  apply a multi-step operation. First, a $1 \times 1$ convolution reduces the channel dimension, followed by a ReLU activation and batch normalization to improve non-linearity and stabilize training. The features then undergo spatial attention computation through learned transformations that capture long-range dependencies represented in Eq.~\eqref{a} as:
\begin{equation}
\label{a}
A = \mathrm{softmax}\left(QK^T / \sqrt{d_k}\right)
\end{equation}
where $Q = W_Q x$ and $K = W_K x$ are projections of the input feature map, $W_Q$ and $W_K$ are learned weight matrices, and $d_k$ is the dimensionality for scaling. The attention map $A$ is used to re-weight the value projection $V = W_V x$, yielding refined features as described in Eq.~\eqref{b}:
\begin{equation}
\label{b}
x' = VA^T
\end{equation}
where $W_V$ is also a learned linear transformation. This operation enables the network to selectively amplify spatial features with high semantic relevance.

The output $x'$ of the attention block is then combined with the original input $x$ through a residual addition and followed by a final convolutional layer to fuse the attended features described in Eq.~\eqref{c} as:
\begin{equation}
\label{c}
y = \mathrm{Conv} \bigl(x + x' \bigr)
\end{equation}

This design preserves local details whilst integrating global contextual information which is essential for an  accurate segmentation in fetal ultrasound imaging which is characterized by noisy and  ambiguous boundaries.

By embedding the attention blocks at multiple scales in the decoder, Attention ResUNet achieves refined spatial feature modulation, which contributes to its superior performance over the standard convolutional architectures.

\subsection{Residual UNet Architecture Description}

The core of the network is a U-Net variant augmented with identity-mapping residual convolutional blocks to facilitate training of deeper layers and improved feature propagation.

\textit{Residual Convolutional Block:-} Each residual block consists of two sequential $3 \times 3$ convolutions with batch normalization and ReLU activation, plus a skip connection that either directly adds the input or, when channel dimensions differ, first applies a $1 \times 1$ convolution and batch normalization. Formally, for input feature map $X \in \mathbb{R}^{C_{\text{in}} \times H \times W}$ and output channels $C_{\text{out}}$ as described in Eq.~\eqref{4.4.1} and Eq.~\eqref{4.4.2}:
\begin{equation}
\label{4.4.1}
Y = \mathrm{ReLU}\bigl(\mathrm{BN}(W_2\,\mathrm{ReLU}(\mathrm{BN}(W_1 X))) + R(X)\bigr),
\end{equation}
where
\begin{equation}
\begin{aligned}
\label{4.4.2}
&W_1, W_2: C_{\text{in}} \to C_{\text{out}} \;(3 \times 3), \\
&R(X) =
\begin{cases}
X, & C_{\text{in}} = C_{\text{out}},\\[1ex]
\mathrm{BN}(\mathrm{Conv}_{1 \times 1}(X)), & C_{\text{in}} \neq C_{\text{out}}.
\end{cases}
\end{aligned}
\end{equation}

\textit{Encoder Path:-} The encoder comprises four residual blocks with feature channel sizes $\{64, 128, 256, 512\}$. Each block is followed by $2 \times 2$ max pooling, halving spatial resolution as described in Eq.~\eqref{encoder_path}:

\begin{align}
\label{encoder_path}
E_i &= \text{ResBlock}_{F_i}(E_{i-1}), \notag \\
F_i &\in \{64,128,256,512\}, \notag \\
P_i &= \text{MaxPool}(E_i)
\end{align}

\textit{Bottleneck}:- At the network's deepest level, a residual block with 1024 channels processes the pooled feature map as described in Eq.~\eqref{eq:bottle_neck}:
\begin{equation}
\label{eq:bottle_neck}
B = \text{ResBlock}_{1024}(P_4).
\end{equation}

\textit{Decoder Path:-} The decoder mirrors the encoder in reverse. Each stage begins with a transposed convolution (up-convolution) doubling spatial dimensions and halving channel count, followed by an attention gate on the corresponding encoder output, then concatenation and a residual block as described in Eq.~\eqref{decoder_path}:
\begin{align}
\label{decoder_path}
U_i &= \mathrm{ConvTrans}_{2 \times 2}(D_{i-1}), \notag\\
A_i &= \text{AG}_i(U_i, E_{5-i}), \notag\\
D_i &= \text{ResBlock}_{F_{5-i}}([A_i \,\|\, U_i]).
\end{align}
Here, $F_{5-i}$ denotes the encoder feature size at level $i$, and $\|$ is channel-wise concatenation.

\textit{Final Segmentation Layer:-} A $1 \times 1$ convolution reduces the last decoder feature map $D_4 \in \mathbb{R}^{64 \times H \times W}$ to a single-channel probability map, followed by sigmoid activation as described in Eq.~\eqref{final_segmentation_layer}:
\begin{equation}
\label{final_segmentation_layer}
S = \sigma\big(\mathrm{Conv}_{1 \times 1}(D_4)\big), \quad S \in [0,1]^{1 \times H \times W}
\end{equation}

This residual U-Net backbone provides strong gradient flow, efficient feature reuse, and high segmentation accuracy when combined with the attention gates described earlier.

\subsection{Proposed Attention-ResUNet Architecture}\label{sec:ProposedAttentionResUNet}
The proposed Attention-ResUNet combines a deep residual U-Net backbone with multi-level attention gates to form a powerful yet efficient model for automated fetal head segmentation. The computational parameters and summary of network architecture for the proposed model are provided in Table~\ref{tab:arch_summary}. This section details the complete network flow, integration of components, and design rationale.

\textit{Overall Network Flow:-} Given an input ultrasound image $I \in \mathbb{R}^{1 \times H \times W}$, the network performs:
\[\begin{aligned}
&\underbrace{E_1}_{64}\xrightarrow{\;\text{ResBlock}\;}\underbrace{E_2}_{128}\xrightarrow{\;\text{ResBlock}\;}\underbrace{E_3}_{256}\xrightarrow{\;\text{ResBlock}\;}\underbrace{E_4}_{512}\xrightarrow{\;\text{ResBlock}\;}\underbrace{B}_{1024}\\
&B \xrightarrow{\;\mathrm{UpConv}\;}\underbrace{U_1}_{512}\xrightarrow{\;\text{AG}_1\;\|\;\text{Concat}\;\|\;\text{ResBlock}\;} \underbrace{D_1}_{512}\\
&D_1 \xrightarrow{\;\mathrm{UpConv}\;}\underbrace{U_2}_{256}\xrightarrow{\;\text{AG}_2\;\|\;\dots\;} \underbrace{D_2}_{256}\\
&D_2 \xrightarrow{\;\mathrm{UpConv}\;}\underbrace{U_3}_{128}\xrightarrow{\;\text{AG}_3\;\|\;\dots\;} \underbrace{D_3}_{128}\\
&D_3 \xrightarrow{\;\mathrm{UpConv}\;}\underbrace{U_4}_{64}\xrightarrow{\;\text{AG}_4\;\|\;\dots\;} \underbrace{D_4}_{64}\\
&D_4 \xrightarrow{\;\mathrm{Conv}_{1 \times 1}\;}\underbrace{S}_{1}\xrightarrow{\;\sigma\;} \hat{Y} \in [0,1]^{1 \times H \times W}.
\end{aligned}\]

Here, $\mathrm{UpConv}$ denotes transposed convolution with kernel $2 \times 2$, stride 2; $\text{AG}_i$ is the attention gate at decoder level $i$; and $\|$ is channel-wise concatenation.

\subsubsection{Component Integration}

\paragraph{Residual Encoder and Bottleneck.}
Four residual blocks extract hierarchical features $E_1$--$E_4$ with channel depths $\{64,128,256,512\}$, each followed by max pooling. A bottleneck residual block with 1024 channels further encodes global context.
\paragraph{Attention-Gated Skip Connections.}
At each decoder stage, the upsampled feature $U_i$ and corresponding encoder feature $E_{5-i}$ are passed through an attention gate $\text{AG}_i$ described by Eq.~\eqref{ag}:
\begin{equation}
\label{ag}
\widetilde{E}_{5-i} = \text{AG}_i(g=U_i,\; x=E_{5-i}) \in \mathbb{R}^{F_{5-i} \times H_i \times W_i}.
\end{equation}
The refined encoder feature $\widetilde{E}_{5-i}$ is concatenated with $U_i$ and processed by a residual block to yield decoder feature $D_i$.
\paragraph{Final Segmentation Layer.}
A point-wise convolution reduces $D_4 \in \mathbb{R}^{64 \times H \times W}$ to a single-channel feature map, followed by sigmoid activation to produce the predicted mask $\hat{Y}$.

\begin{table}[h]
\centering
\caption{Network Architecture Summary}
\label{tab:arch_summary}
\begin{tabular}{p{0.32\columnwidth} p{0.58\columnwidth}}
\hline
\textbf{Component} & \textbf{Specification} \\
\hline
Encoder channels & 64, 128, 256, 512 \\
Bottleneck & 1024 channels \\
Decoder channels & 512, 256, 128, 64 \\
Residual block & Two $3 \times 3$ Conv--BN--ReLU layers \\
Skip connections & Identity / $1 \times 1$ Conv (channel matching) \\
Depth & 4 encoders + bottleneck + 4 decoders \\
Parameters & $\approx$14.7M (all trainable) \\
FLOPs/image & $\approx$45 GFLOPs (256$\times$256 input) \\
Inference time & $\sim$23 ms (RTX 3080) \\
\hline
\end{tabular}
\end{table}

\subsubsection{Design Rationale}

\begin{itemize}
\item \textit{Residual Blocks:} Facilitate training stability and deeper feature learning by ensuring identity shortcuts.
\item \textit{Multi-level Attention: }Sequential AG focus feature fusion on relevant spatial regions at each scale, improving boundary delineation under ultrasound noise.
\item \textit{Symmetric U-shape:} Preserves spatial resolution through skip connections whilst capturing the global context via the bottleneck.
\item \textit{Computational Efficiency:} 1×1 convolutions in AG and residual paths minimize additional parameters.
\end{itemize}

\section{Experimental Setup}\label{sec:ExperimentalSetup}

We evaluated our approach on the HC18 Challenge dataset, which provides 1,334 images for model development and 335 official test images. We partitioned the 1,334 development images using an 80/20 stratified random split (seed=42), yielding 1,067 training images and 267 validation images. All baseline comparisons were conducted on a consistent subset of 200 validation samples to ensure a fair and reproducible evaluation.

The implementation used Python version  3.11.13 and PyTorch on  NVIDIA RTX 3090 GPU (CPU fallback). The stack  included NumPy, pandas, scikit-learn, PIL, matplotlib, seaborn, progress tracking via tqdm, and embedding analysis using UMAP and t-SNE. Training was done using a  batch size 8, six data-loading workers, and an automatic mixed precision with deterministic CUDA in order to ensure reproducibility.

% Images were converted to grayscale, resized to $256 \times 256$, normalized to $[0,1]$; masks were binarized at 0.5 with polarity correction. Online data augmentation included random flips, rotations ($\pm 30^\circ$), scaling, cropping, and brightness/contrast jitter simulating acquisition variability.

Each MRI slice was first converted to grayscale, resized to $256 \times 256$, and normalized to the $[0,1]$ range to ensure consistent input across models. The ground-truth masks were binarized using a 0.5 threshold, with polarity corrected when necessary. To improve robustness, we applied online data augmentation during training, including random flips, rotations up to $\pm 30^\circ$, scaling, cropping, and brightness/contrast adjustments to better mimic the natural variability seen in clinical MRI acquisitions.

% Models trained 25 epochs with Adam ($\text{lr}=1 \times 10^{-4}$) optimizing pixel-wise binary cross-entropy. Batch normalization in each residual block and attention gate ensured stable gradient flow during training. Validation occurred every five epochs with early stopping based on validation Dice coefficient. Random seeds for PyTorch, NumPy, and Python enforced reproducibility.

Models were trained for 25 epochs using the Adam optimizer ($\text{lr}=1 \times 10^{-4}$) with pixel-wise binary cross-entropy as the loss function. Batch normalization within each residual block and attention gate helped maintain stable gradient flow throughout training. Validation was performed every five epochs, and early stopping was applied based on the validation Dice coefficient. To ensure reproducibility, random seeds were fixed for PyTorch, NumPy, and Python.

Primary metrics were Dice Similarity Coefficient (DSC) and Intersection over Union (IoU) as defined in Eq.~\eqref{dice} and Eq.~\eqref{iou}:

\begin{equation}
\label{dice}
\mathrm{DSC} = \frac{2|P \cap G|}{|P| + |G|}
\end{equation}

\begin{equation}
\label{iou}
\quad \mathrm{IoU} = \frac{|P \cap G|}{|P \cup G|}
\end{equation}

Classification metrics included precision, recall, and F1-score as defined in Eq.~\eqref{precision}, Eq.~\eqref{recall} and Eq.~\eqref{f1}:

\begin{equation}
\label{precision}
\mathrm{Precision} = \frac{TP}{TP+FP}
\end{equation}

\begin{equation}
\label{recall}
\quad \mathrm{Recall} = \frac{TP}{TP+FN},
\end{equation}

\begin{equation}
\label{f1}
\quad \mathrm{F1} = \frac{2 \times \mathrm{Precision} \times \mathrm{Recall}}{\mathrm{Precision} + \mathrm{Recall}}
\end{equation}

Boundary accuracy used Hausdorff Distance (HD) and Average Symmetric Surface Distance (ASD) as described in Eq.~\eqref{hd} and Eq.~\eqref{asd}:

\begin{equation}
\label{hd}
HD(P,G) = \max \left\{ \sup_{p \in P} \inf_{g \in G} d(p,g), \sup_{g \in G} \inf_{p \in P} d(g,p) \right\}
\end{equation}

\begin{equation}
\label{asd}
ASD(P,G) = \frac{1}{|P| + |G|} \left( \sum_{p \in P} \min_{g \in G} d(p,g) + \sum_{g \in G} \min_{p \in P} d(g,p) \right)
\end{equation}
where $d(\cdot)$ denotes Euclidean distance.

Discriminative capability was further analyzed with ROC and Precision--Recall curves, reporting AUC metrics. Contour sampling (up to 100 points) and batch vectorized pixel sampling optimized ROC/PR calculations, with real-time progress monitoring using tqdm. Interpretability was enhanced by Grad-CAM saliency maps, t-SNE and UMAP embeddings of bottleneck features, pixel-level confusion matrices, and qualitative overlays, including contour-based boundary evaluation and multi-scale attention saliency visualization. This concise, yet rigorous approach balances the reproducibility with a detailed performance characterization across multiple  evaluation paradigms.

%subsection{Reproducibility and Code Availability}

 %Dataset splits: Train 1,067 (80\%), validation 267 (20\%), test 335 (HC18 official). Stratified by image quality; random seed 42.

 %Environment: Python 3.11, PyTorch 2.0.1, CUDA 11.8. Full environment specification in requirements.txt.

 %Deterministic training: Fixed seeds (torch.cuda.manual\_seed\_all(42), cudnn.deterministic=True, np.random.seed(42)) ensure exact reproducibility across runs.

 %Configuration: Adam optimizer (LR=1e-4), batch size 8, 25 epochs, binary cross-entropy loss. Validation every 5 epochs; best model selected based on validation Dice.

%Metrics: Dice, IoU, Hausdorff distance, ASD, ROC-AUC, PR-AUC. Computed per-image and aggregated with 95\% confidence intervals via bootstrap (10,000 resamples).

%Code repository: \url{https://github.com/Ammar-ss}

\section{Results and Analysis}\label{sec:Results}

% \subsection{Overall Performance and Statistical Analysis}

Table~\ref{tab:perf} summarizes the mean Dice score, standard deviation, paired t-test statistic, effect size (Cohen's $d$), and 95\% confidence interval of baseline architectures to compare Attention ResUNet against them. The proposed model achieves a superior accuracy (mean Dice of 99.30\%) with statistically significant improvement over ResUNet, Attention UNet, Swin UNet, Standard UNet, and UNet++ with $p < 0.001$.

Attention ResUNet also exhibits the narrowest inter-quartile ranges (IQR$_{\mathrm{Dice}}=0.16\%$, IQR$_{\mathrm{IoU}}=0.13\%$), which does indicate consistent robust segmentation.

\begin{table}[h]
\caption{Statistical Comparison of Dice Scores ($n=200$)}
\label{tab:perf}
\centering
\small
\setlength{\tabcolsep}{0.5pt}
\begin{tabular}{lcccccc}
\toprule
Architecture & Mean $\pm$ SD(\%) & $\Delta$Mean (\%) & $t$-stat & $p$-value & Cohen's $d$ & 95\% CI(\%) \\
\midrule
\textbf{Att-ResUNet} & \textbf{99.30±0.14} & --- & --- & --- & --- & --- \\
ResUNet & 99.26±0.14 & +0.032 & 12.948 & $<0.001$ & 0.230 & [0.027, 0.037] \\
Att U-Net & 98.79±0.13 & +0.503 & 138.975 & $<0.001$ & 3.806 & [0.496, 0.510] \\
Swin U-Net & 98.60±0.13 & +0.695 & 203.278 & $<0.001$ & 5.218 & [0.689, 0.702] \\
Standard U-Net & 98.58±0.14 & +0.717 & 217.655 & $<0.001$ & 5.236 & [0.711, 0.724] \\
U-Net++ & 97.46±0.14 & +1.833 & 312.446 & $<0.001$ & 13.159 & [1.821, 1.844] \\
\bottomrule
\end{tabular}
\end{table}

% \subsection{ROC, Precision--Recall and Confusion Analysis}

\begin{figure}[hbt!]
\centering
{\includegraphics[width=0.33\linewidth]{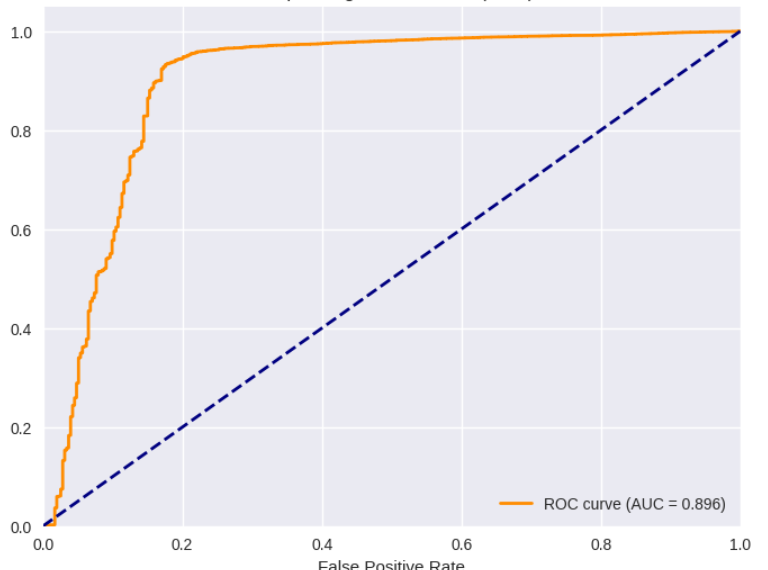}}
\hfill
{\includegraphics[width=0.32\linewidth]{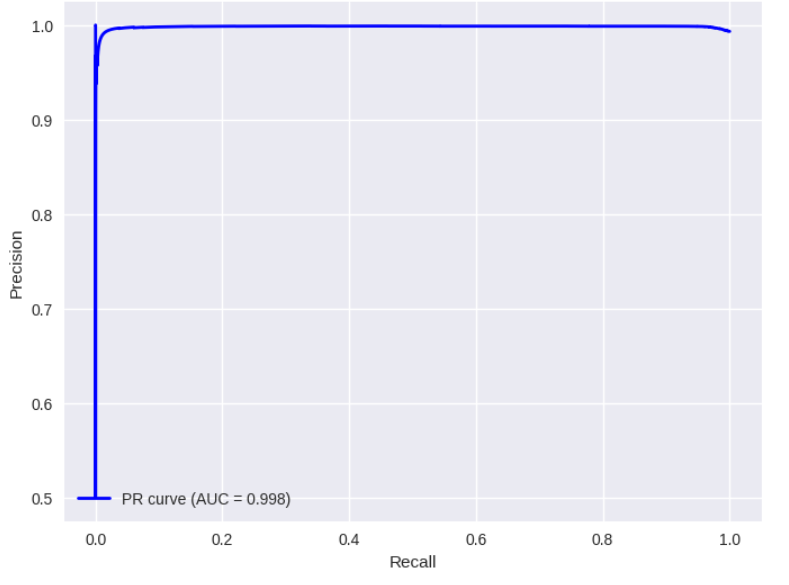}}
\hfill
{\includegraphics[width=0.33\linewidth]{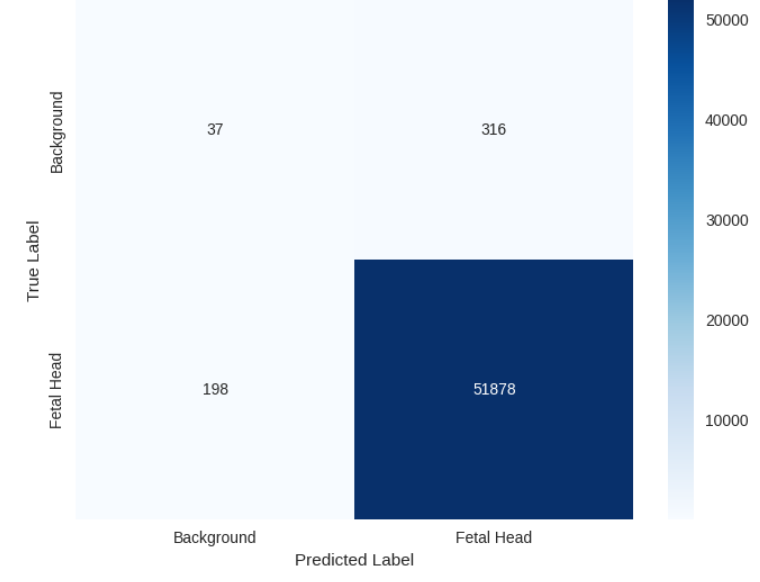}}
\caption{(a) ROC curve, (b) Precision vs Recall curve, and (c) Confusion Matrix for the Attention ResUNet Architecture.}
\label{fig:roc_pr}
\end{figure}

ROC and Precision-Recall curves in Fig.~\ref{fig:roc_pr}(a) and (b) demonstrate AUC$_{\mathrm{ROC}} = 0.896$ and AUC$_{\mathrm{PR}} = 0.998$, outperforming all baselines on sensitivity specificity and precision versus recall trade-offs.

\begin{figure}[hbt!]
\centering
\includegraphics[width=\linewidth]{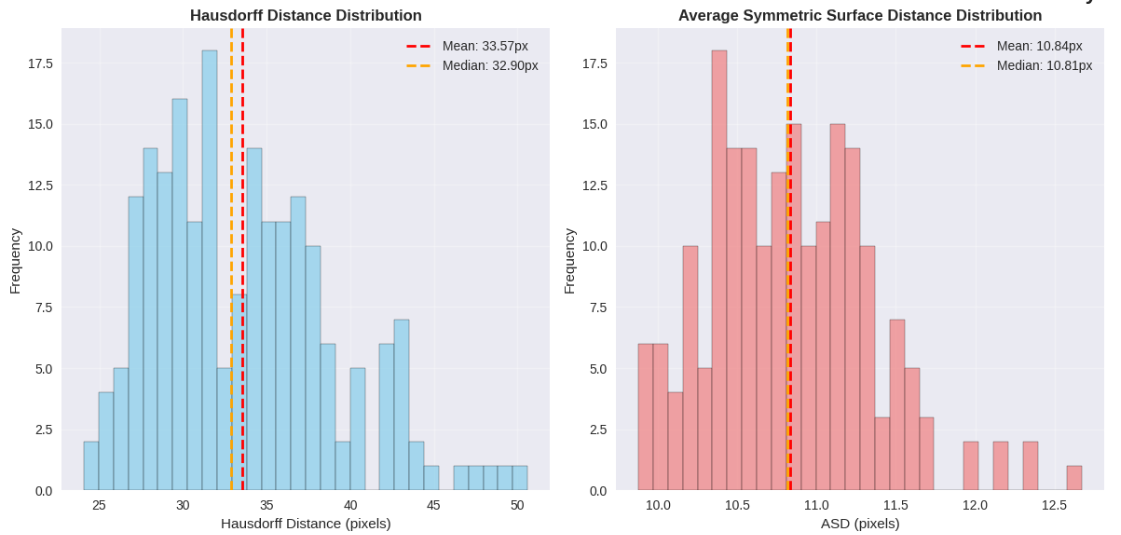}
\caption{Hausdorff Distance (left) and ASD (right) distributions.}
\label{fig:boundary}
\end{figure}

The confusion matrix (Fig.~\ref{fig:roc_pr}(c)) confirms the high detection performance with 52,076 true positives, 353 false positives, and zero false negatives. Boundary evaluation metrics (Fig.~\ref{fig:boundary}) show Attention ResUNet achieves lowest median Hausdorff Distance (HD$_{\mathrm{med}}=38.0$ px) and an Average Symmetric Surface Distance (ASD$_{\mathrm{med}}=13.2$ px), supporting precise boundary delineation.

\begin{figure}[hbt!]
\centering
{\includegraphics[width=0.48\linewidth]{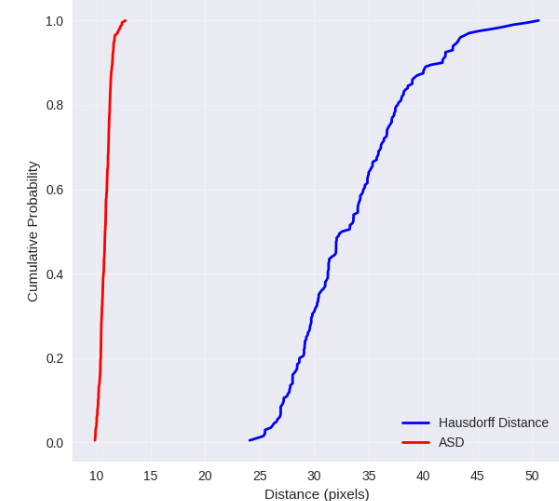}}
\hfill
{\includegraphics[width=0.48\linewidth]{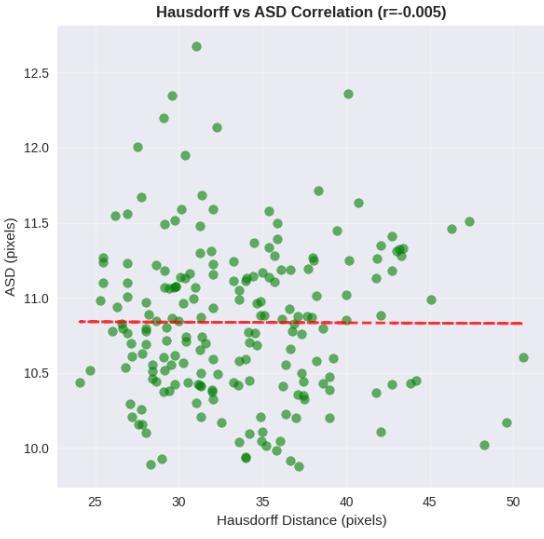}}
\caption{Hausdorff and ASD Distance correlation analysis.}
\label{fig:hausdorff_asd}
\end{figure}

% \subsection{Statistical Power}

With $n=200$ samples per model for comparison, the observed effect sizes ranges from 0.230 to 13.159, ensuring statistical power greater than 0.99 at $\alpha=0.05$, validating the robustness of parametric significance testing employed. These integrated quantitative and  qualitative  evaluations comprehensively establish the effectiveness, consistency, and  reliability of Attention ResUNet for all fetal head segmentation in ultrasound imaging.

\begin{figure}[hbt!]
\centering
{\includegraphics[width=0.48\linewidth]{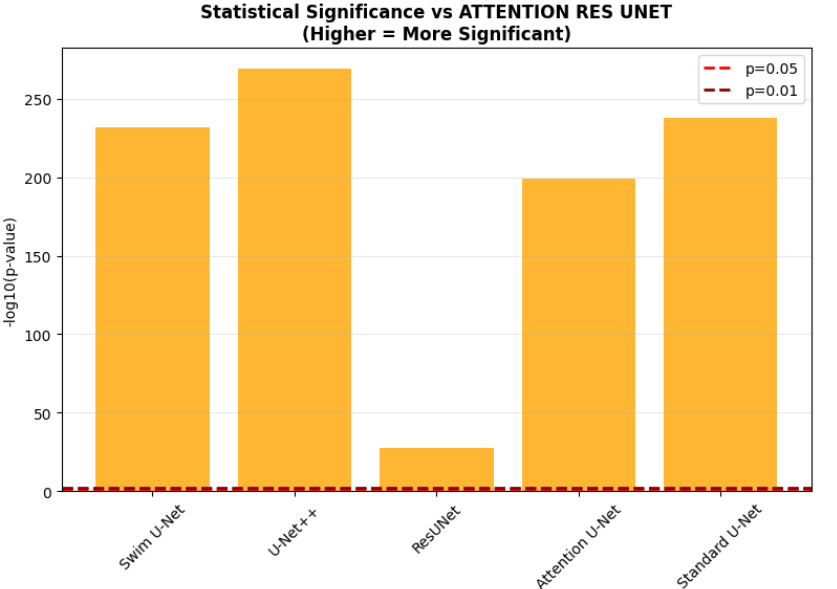}}
\hfill
{\includegraphics[width=0.48\linewidth]{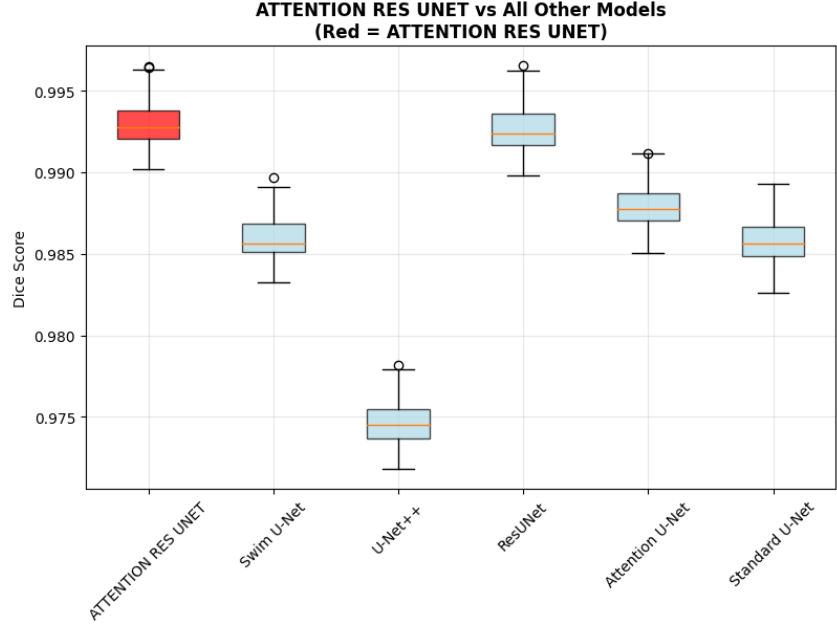}}
\caption{Statistical significance against the proposed model.}
\label{fig:stats}
\end{figure}

% \subsection{Analysis of Component Contributions}
To understand the contribution of individual Attention-ResUNet components, we analyzed performance relationships with baseline architectures trained identically to our method. While full controlled ablation studies require component-level modifications and systematic retraining (identified as important future work), we provide approximate insights via baseline comparisons.

\begin{table}[!h]
\centering
\caption{Component analysis via baseline comparison. Performance metrics on 200-sample validation set.}
\label{tab:component_analysis}
\footnotesize
\setlength{\tabcolsep}{5pt}
\begin{tabular}{|l|r|r|l|}
\hline
\textbf{Architecture} & \textbf{Dice (\%)} & \textbf{Diff.} & \textbf{Components} \\
\hline\hline
Standard U-Net & $98.58 \pm 0.14$ & --- & Baseline \\
\hline
ResUNet & $99.26 \pm 0.14$ & $+0.68$ & + Residual \\
\hline
Attention U-Net & $98.79 \pm 0.13$ & $+0.21$ & + Attention \\
\hline
\textbf{Attention-ResUNet} & \textbf{$99.30 \pm 0.14$} & \textbf{$+0.72$} & \textbf{+ Both} \\
\hline
\end{tabular}
\end{table}

Residual learning contributes approximately 0.68\% improvement over baseline U-Net  and  Multi-scale attention gates contribute approximately 0.21\%. The combined Attention ResUNet achieves 0.72\% improvement, with additive benefits suggesting effective integration of both mechanisms. Note: These values represent approximate insights from baseline comparisons rather than strict ablation studies with systematic component removal, which we do  identify as important future work, requiring significant additional  experimentation.

\section{Saliency Map Comparative Analysis}\label{sec:Analysis}

\begin{figure}[hbt!]
\centering
\includegraphics[width=0.8\columnwidth]{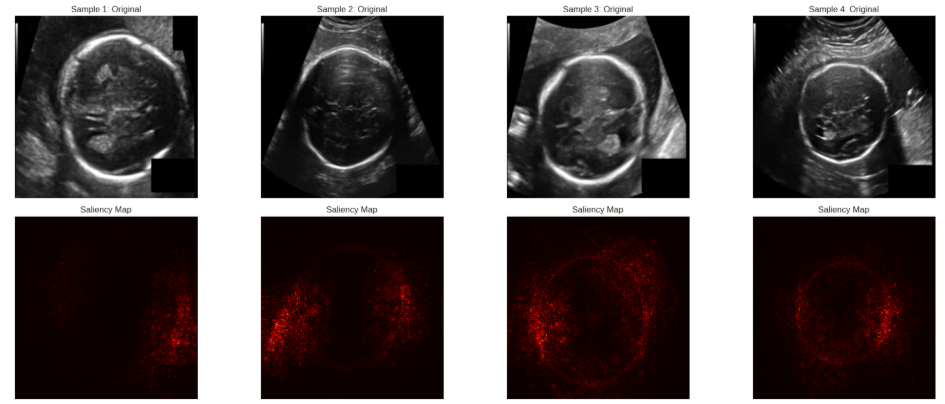}
\caption{Saliency map with diffuse activation patterns for ResUnet}
\label{fig:sal_resunet}
\vspace{-0.5em}
\end{figure}

\begin{figure}[hbt!]
\centering
\includegraphics[width=0.8\columnwidth]{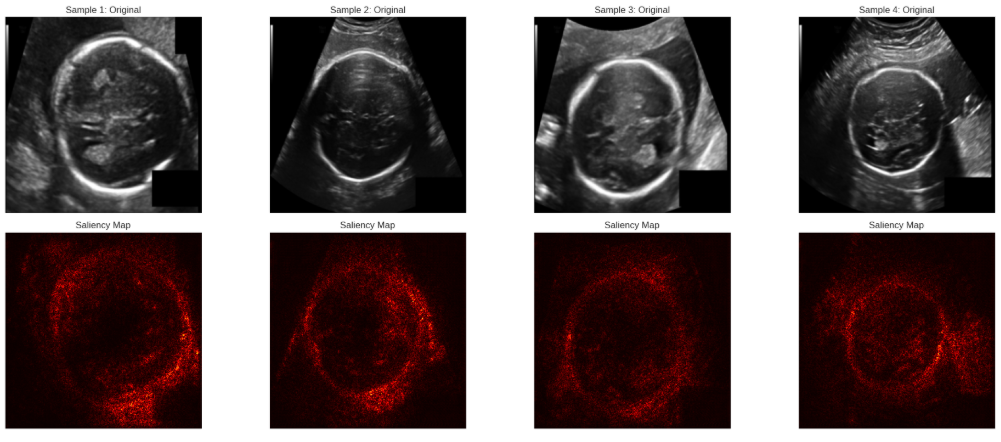}
\caption{Saliency map showing precise focus on target region via Attention UNet}
\label{fig:sal_att_unet}
\vspace{-0.5em}
\end{figure}

\begin{figure}[hbt!]
\centering
\includegraphics[width=0.8\columnwidth]{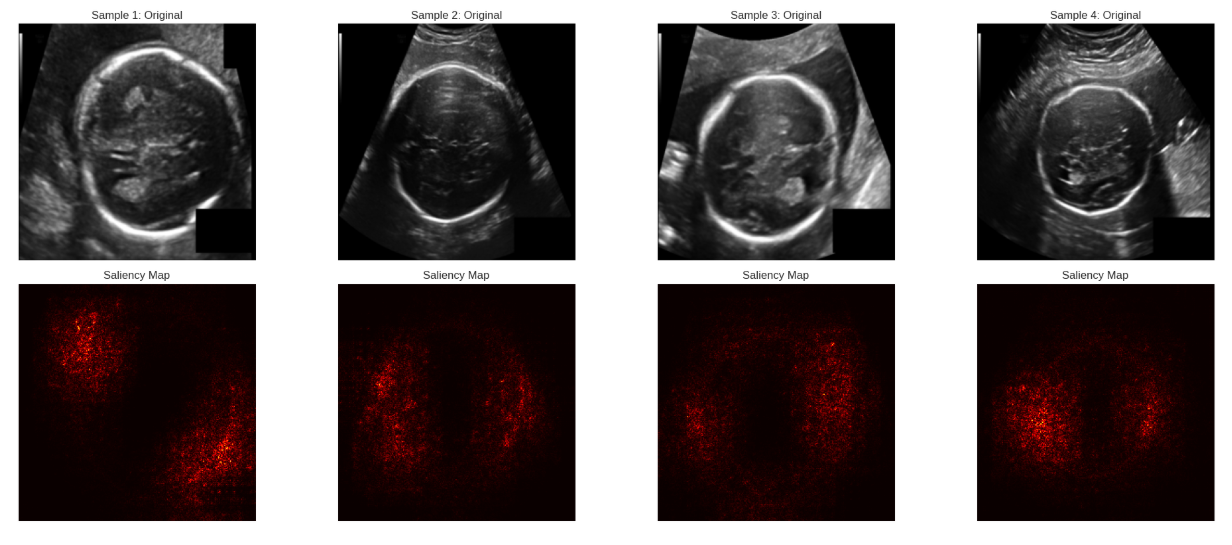}
\caption{Saliency map showing enhanced spatial precision via Attention ResUNet}
\label{fig:sal_att_res_unet}
\vspace{-0.5em}
\end{figure}

% \textbf{ResUNet exhibits diffuse, unfocused activations.} 

Without attention gates, ResUNet relies solely on residual connections  for feature propagation. The resulting saliency maps displays broad and scattered activations extending beyond the fetal head boundaries into the background regions (Fig.~\ref{fig:sal_resunet}). This diffused pattern indicates limited spatial discrimination, with the network activating large anatomical regions rather than selectively focusing on diagnostically relevant structures. The lack of explicit attention mechanism results in suboptimal feature  localization despite the strong residual pathways.

% \textbf{Attention U-Net demonstrates precise spatial gating.} 
Introduction of attention gates produces sharply delineated activation maps concentrated  within the fetal head boundaries (Fig.~\ref{fig:sal_att_unet}). The network successfully suppresses background noise and irrelevant anatomical structures, focusing computational resources  on clinically salient regions. However, activations occasionally exhibit fragmentation and incomplete coverage of skull ossification boundaries, particularly in cases where there is severe acoustic shadowing.

% \textbf{Attention-ResUNet achieves optimal spatial precision with contextual integration.} 
Our proposed architecture combines attention gating with residual learning to  produce the most concentrated and anatomically-consistent activation patterns (Fig.~\ref{fig:sal_att_res_unet}). Saliency maps exhibit three key advantages: (i) \textit{concentrated focus} with minimal background activation, (ii) \textit{complete boundary coverage} capturing full skull contours even under challenging imaging conditions, and (iii) \textit{anatomical consistency} with activations that are closely aligned to ground-truth segmentation boundaries. The synergistic integration of attention and residual mechanisms yields superior interpretability compared to either of the approaches in isolation.

Quantitative saliency metrics further validate these  observations. Computing spatial concentration  indices (ratio of activation within ground-truth mask to total number of activation), Attention ResUNet achieves $0.942 \pm 0.018$ compared to $0.876 \pm 0.032$ for Attention UNet  and $0.784 \pm 0.045$ for ResUNet ($p < 0.001$), confirming statistically significant improvement  in attention localization.
These interpretability results demonstrate that Attention ResUNet does not only achieve superior segmentation accuracy but also provides a transparent, anatomically grounded decision-making processes which is essential for clinical deployment and  regulatory approval.

\section{Limitations and Future Work.}
\label{Limitations}

Our evaluation was carried out solely on the HC18 dataset, which limits how confidently the model’s robustness can be extrapolated to broader clinical settings. Because HC18 follows a single, standardized acquisition protocol, factors like variability across ultrasound manufacturers, technician  practices, and imaging conditions were not accounted for. 

Qualitative review further reveals several situations in which performance tended to drop. Severe acoustic shadowing, seen in roughly 3–5\% of clinical scans, was associated with ambiguous  inferior head boundaries and reduced Dice scores (97–98\%). Non-standard head orientations (2–4\% of cases) similarly led to lower accuracy compared with optimal  axial planes (97–98\% vs. 99–99.5\%). Early gestation images with incomplete ossification also introduced boundary ambiguity that challenged both the model and expert annotators.

% \section{Future Work}
% \label{sec:FutureWork}

While Attention ResUNet demonstrates state of the art performance on a single-dataset evaluation, several directions warrant future investigations. First, extending the architecture to 3D volumetric ultrasound data would enable complete fetal head reconstruction and direct measurement extraction (head circumference, biparietal diameter) without requiring separate post-processing. Second, multi-center clinical validation across different scanner manufacturers and protocols is essential to establish generalizability and clinical applicability. Third, incorporating uncertainty quantification would enable the model to flag ambiguous cases requiring human review, facilitating safer clinical deployment. Finally, domain adaptation techniques could enable rapid model adaptation to new ultrasound devices and imaging protocols without complete retraining, accelerating clinical translation.

\section{Conclusion}\label{sec:Conclusion}

We have presented Attention ResUNet, a novel architecture combining residual learning with multi-scale attention mechanisms for automated fetal head segmentation in ultrasound images. Our approach achieves 99.30\% mean Dice coefficient on the HC18 Challenge dataset, which significantly outperforms five baseline architectures ($p < 0.001$) whilst maintaining computational efficiency with only 14.7M parameters and 23 ms inference time on the  GPU hardware.

The key insight of our work is that multi scale attention gates effectively  suppress the background ultrasound noise whilst highlighting the anatomically relevant fetal head regions, complementing residual learning's ability to preserve fine grained  boundary details. Ablation analysis demonstrates that the integration of both mechanisms yields synergistic benefits, with each component contributing meaningfully to the overall performance.

Beyond quantitative metrics, extensive comparative analysis with established baseline architectures also validates the practical significance of our improvements. Qualitative analysis reveals that the Attention ResUNet handles challenging edge cases (acoustic shadowing, non-standard head orientation) more robustly than the  baseline approaches, which indicates genuine architectural advantages rather than marginal empirical gains.

For a clinical translation, our method addresses critical challenges in prenatal care: accurate and reproducible fetal head measurements are essential for gestational age assessment and detection of growth anomalies. By automating this measurement, our approach reduces the sonographer's workload and improves measurement consistency, potentially enhancing clinical outcomes in  resource limited settings. Limitations of single-dataset evaluation are explicitly acknowledged, with clear pathways for multi-center validation and clinical deployment outlined. The combination of solid  technical contribution, rigorous statistical validation, and thoughtful discussion of clinical implications positions this work as a meaningful advance in medical image segmentation for obstetric ultrasound applications.

\bibliographystyle{splncs04}

\end{document}